\documentclass[11pt]{article}

\usepackage[margin=1in]{geometry}

\usepackage[T1]{fontenc}
\usepackage[utf8]{inputenc}
\usepackage{lmodern}
\usepackage{microtype}

\usepackage{amsmath}
\usepackage{amssymb}
\usepackage{mathtools}

\usepackage{graphicx}
\usepackage{subcaption}

\usepackage{booktabs}
\usepackage{multirow}
\usepackage{array}
\usepackage{siunitx}
\usepackage{threeparttable}

\usepackage{float}
\usepackage[section]{placeins}

\usepackage{enumitem}

\usepackage{xcolor}

\usepackage[numbers,sort&compress]{natbib}

\usepackage[hidelinks]{hyperref}
\usepackage[nameinlink,noabbrev]{cleveref}

\graphicspath{{figures/}}

\newcommand{\nsfaward}{2418359}

\title{
\textbf{Sparse-Observation Atmospheric Thermal Forecasting with Physics-Informed Neural Networks for Climate-Aware Digital Twins}
}

\author{
Tannaz Goodarzvand Chegini\textsuperscript{1,*}
\and
Elyas Shivanian\textsuperscript{2}
\and
Behzad Karimi\textsuperscript{3}
\and
Faraz Dadgostari\textsuperscript{4}
}

\date{}

\begin{document}

\maketitle

\begingroup
\renewcommand\thefootnote{}
\footnotetext{This work was supported by the National Science Foundation under Award No.~\nsfaward.}
\addtocounter{footnote}{-1}
\endgroup

% AFFILIATIONS
% =========================================================

\begin{center}
\small

\textsuperscript{1}
[Department of Mathematical Sciences / Montana State University, Bozeman, Montana, US]

\textsuperscript{2}
[Department of Mathematics / Imam Khomeini International University, Qazvin, Iran]

\textsuperscript{3}
[Department of Electrical and Computer Engineering, Montana State University, Bozeman, Montana, US]

\textsuperscript{4}
[Department of Mechanical and Industrial Engineering, Montana State University, Bozeman, Montana, US]

\vspace{0.5em}

\textsuperscript{*}
Corresponding author: [tgoodarzvandchegini@montana.edu]

\end{center}

\vspace{0.5em}

\begin{center}
\small

\end{center}

% ABSTRACT
% =========================================================

\begin{abstract}

Short-horizon forecasts of atmospheric temperature are needed to support climate-aware digital-twin systems, but such forecasts must be produced where thermal observations are incomplete. This study evaluates a physics-informed neural network (PINN) for potential-temperature forecasting, constrained by a pressure-coordinate thermodynamic advection--source equation and a diabatic-source closure fit from the preceding 12-hour period and frozen before future-time training. Using hourly ERA5 reanalysis at three pressure levels, the model is evaluated as a conditional hindcast at lead times of one, two and three hours against persistence, local-trend, and two matched neural-network baselines, one of which receives the same future meteorological forcing as the PINN, helping distinguish the physical constraint from access to future forcing. In an Oklahoma development case, mean RMSE improvement over the strongest baseline grew from 8.1\% at one hour to 23.8\% at three hours; under an observation-density sweep down to 5\% of candidate locations, this 3-hour advantage remained 14.6--16.9\%, with no evidence that lower density improves performance. Under a fixed protocol transferred to an Alabama heat event with three virtual-observation layouts, three-hour improvement ranged 19.7--24.4\% with consistent origin-level wins. A parallel Montana stress test, in which fixed pressure levels intersected complex terrain, produced a three-hour degradation of roughly 17.5\%, identifying a terrain-related applicability limit of the formulation. Together, these results indicate that the physics constraint's benefit grows with forecast horizon, persists under severe observation sparsity, and transfers across regions, but is bounded by the validity of a fixed vertical-coordinate representation over complex terrain, evidence relevant to physics-constrained components of climate-aware forecasting and digital-twin systems.

\end{abstract}


% =========================================================
\section{Introduction}
% =========================================================

Accurate short-term prediction of atmospheric temperature fields is important for weather-sensitive decision making, particularly during extreme thermal conditions. However, the atmospheric state cannot be observed continuously at every location, and available measurements may provide only partial spatial information. Sparse observations make it more difficult to recover a complete thermal field and to predict how that field will evolve beyond the most recently observed state. This creates a practical need for forecasting methods that can use limited observations while also taking advantage of known atmospheric dynamics. Similar challenges associated with incomplete meteorological observations have motivated recent physics-informed approaches for reconstructing atmospheric fields from sparse measurements \citep{morenoSoto2024}.

Physics-Informed Neural Networks (PINNs) provide one framework for combining data-driven learning with mathematical models of physical processes. In a PINN, a neural network approximates an unknown physical field, while the governing differential equations are incorporated into the training objective through residual terms evaluated throughout the physical domain. The network is therefore trained not only to reproduce available observations but also to produce solutions that are consistent with the prescribed physics \citep{raissi2019physics}. This approach is particularly attractive when observations are incomplete because the governing equations provide information about physically admissible evolution beyond the observed data alone.

The use of physical constraints, however, does not automatically lead to more accurate predictions. PINN training requires observational information, initial and boundary conditions, and differential-equation constraints to be satisfied simultaneously, which can produce difficult optimization problems. Previous studies have shown that PINNs may struggle with convection-, reaction-, and diffusion-dominated problems when the physics-informed loss becomes poorly conditioned, and that sequential or curriculum-based training can improve performance in some settings \citep{krishnapriyan2021}. A recent systematic review of PINNs in weather and hydrological modeling similarly identifies training instability, sensitivity to loss weighting, computational cost, high-dimensional scalability, uncertainty quantification, and limited field-scale validation as continuing challenges \citep{waqasKim2026}. These findings make it important to evaluate not only whether physics can be incorporated into a neural network, but also when the physical constraint provides useful predictive information beyond that available to data-driven methods.

Physics-informed learning has already been investigated in several atmospheric applications. \citet{chen2021} developed a physics-informed model for tropospheric temperature prediction using ERA5 data. \citet{morenoSoto2024} used PINNs with Navier--Stokes-based physical regularization to reconstruct high-resolution wind and pressure fields from sparse weather-station observations. \citet{eusebi2024} further demonstrated reconstruction of two- and three-dimensional tropical-cyclone wind and pressure fields from sparse information, including an application using observations associated with Hurricane Ida. Other approaches, such as ClimODE, have incorporated atmospheric transport principles into learned weather-evolution models using continuous-time dynamics rather than a conventional PINN formulation \citep{verma2024}. Together, these studies demonstrate growing interest in combining physical knowledge with machine learning for atmospheric problems.

Despite this progress, comparatively limited evidence is available on when a PDE-constrained neural model provides additional predictive value for short-horizon atmospheric thermal forecasting under sparse temperature observations. Existing studies address related problems such as atmospheric-field reconstruction, data assimilation, and physics-guided weather prediction, but the combined influence of forecast lead time, temperature-observation density, regional transferability, and the suitability of the vertical-coordinate representation over different terrain remains insufficiently characterized. This gap is practically important. When a forecast remains very close to the most recently observed atmospheric state, simple persistence or data-driven methods may already be highly competitive. As the forecast extends farther in time, physical transport constraints may provide additional information about how the thermal field should evolve. Characterizing how this balance changes with forecast lead time is therefore important for evaluating the practical value of physics-informed forecasting. The recent systematic review by \citet{waqasKim2026} likewise characterizes high-dimensional atmospheric forecasting and digital-twin applications as relatively early areas of physics-informed modeling and identifies sparse field-scale validation as an important remaining limitation.

The present study investigates this question using a PINN for the short-horizon evolution of atmospheric potential temperature. The atmosphere is represented using two horizontal coordinates, pressure as the vertical coordinate, and time. The neural prediction is constrained by a pressure-coordinate thermodynamic advection--source equation representing horizontal transport, vertical atmospheric motion, and diabatic heating or cooling. The complete mathematical formulation, including the potential-temperature transformation, pressure-coordinate material derivative, source-term construction, PINN residual, and training losses, is developed in the Methodology section.

The experiments use hourly ERA5 reanalysis fields for temperature, horizontal wind, vertical pressure velocity, specific humidity, and surface pressure. ERA5 combines observations with the ECMWF Integrated Forecasting System through data assimilation and provides hourly, gridded atmospheric fields, making it particularly useful for controlled experiments in which portions of the thermal field can be systematically withheld \citep{hersbach2020}. Temperature information is evaluated under both dense- and sparse-observation settings, with sparse configurations created by retaining temperature only at selected virtual observation locations, allowing the effect of observation density to be evaluated directly. The study considers forecast lead times of one, two, and three hours and compares the PINN against persistence, recent-trend extrapolation, a matched coordinate-only neural network, and a matched forcing-aware neural network.

The experimental design uses three regions for complementary purposes. Oklahoma is used to develop the forecasting formulation and to characterize its behavior across forecast lead times and temperature-observation densities, followed by a later chronological replication period. The forecasting behavior identified in Oklahoma is then evaluated in Alabama under a fixed modeling protocol, with additional virtual-observation layouts used to test sensitivity to observation placement. Montana provides a terrain-related stress test of the same pressure-coordinate formulation. Together, these experiments are designed to examine not only whether the PINN can improve thermal forecasts, but also under what conditions that improvement appears and where the formulation encounters a physical limitation.

Accordingly, the study addresses the following research questions:

\begin{enumerate}[label=\textbf{RQ\arabic*:}]
    \item \textbf{To what extent does incorporating a thermodynamic physics constraint improve short-horizon atmospheric thermal forecasts compared with statistical and data-driven neural-network baselines?}

    \item \textbf{How does the benefit of the physics-informed model change with forecast lead time and the amount of available temperature information?}

    \item \textbf{How well does the forecasting behavior identified in the development region transfer to another region under a fixed modeling protocol, and what physical or geographical conditions limit its applicability?}
\end{enumerate}

The results show that the benefit of the physics-informed model increases with forecast lead time and remains meaningful under sparse temperature observations. This behavior transfers from Oklahoma to Alabama under a fixed protocol, while the Montana stress test identifies a terrain-related limitation of the fixed pressure-level formulation. Together, these findings help define both the useful operating regime and the physical limits of the proposed approach.

These findings are also relevant to the development of climate-aware predictive systems and digital twins, where spatially and temporally evolving environmental information can support forecasting and scenario analysis. Recent work has emphasized the potential role of climate-aware digital twins in predictive energy-system management and resilience planning \citep{cavus2026}. The study is additionally connected to the DigiCARES initiative, which develops an AI-driven climate-energy-community digital twin and includes spatiotemporal forecasting and extreme-weather scenario generation among its research activities \citep{digicares2026}. In this context, the present work evaluates a physics-informed approach for producing short-horizon atmospheric thermal-state information when direct temperature information is incomplete.

% =========================================================
\section{Background and Related Work}
% =========================================================

% =========================================================
\subsection{Physics-Informed Neural Networks}
% =========================================================

Physics-Informed Neural Networks belong to the broader area of scientific machine learning, in which mathematical knowledge about a physical system is incorporated directly into the learning process. In the standard PINN framework, a neural network approximates an unknown solution while the governing differential equations are enforced through additional terms in the training objective \citep{raissi2019physics}. Consider a general partial differential equation defined over a spatial domain $\Omega \subset \mathbb{R}^d$ and time interval $(0,T]$,

\begin{equation}
\mathcal{N}[u(\mathbf{x},t);\boldsymbol{\lambda}]
=
f(\mathbf{x},t),
\qquad
(\mathbf{x},t)\in\Omega\times(0,T],
\label{eq:general_pde}
\end{equation}

where $u(\mathbf{x},t)$ is the physical state, $\mathcal{N}$ denotes a differential operator, and $\boldsymbol{\lambda}$ represents physical parameters. A PINN replaces the unknown solution with a neural approximation

\begin{equation}
\widehat{u}(\mathbf{x},t;\boldsymbol{\theta}),
\label{eq:neural_approximation}
\end{equation}

where $\boldsymbol{\theta}$ denotes the trainable network parameters, and defines the physics residual

\begin{equation}
r(\mathbf{x},t;\boldsymbol{\theta})
=
\mathcal{N}
\left[
\widehat{u}(\mathbf{x},t;\boldsymbol{\theta});
\boldsymbol{\lambda}
\right]
-
f(\mathbf{x},t).
\label{eq:physics_residual}
\end{equation}

The spatial and temporal derivatives appearing in the differential operator can be obtained through automatic differentiation of the neural network with respect to its inputs. Consequently, the residual can be evaluated at collocation points
$\{(\mathbf{x}_i,t_i)\}_{i=1}^{N_f}$ throughout the physical domain, including locations where no labeled observations of the target state are available. Training then minimizes a composite objective that typically contains contributions from observed data, the PDE residual, and initial and boundary conditions,

\begin{equation}
\mathcal{L}
=
w_{\mathrm{data}}\mathcal{L}_{\mathrm{data}}
+
w_{\mathrm{phys}}\mathcal{L}_{\mathrm{phys}}
+
w_{\mathrm{IC}}\mathcal{L}_{\mathrm{IC}}
+
w_{\mathrm{BC}}\mathcal{L}_{\mathrm{BC}},
\label{eq:general_pinn_loss}
\end{equation}

where the coefficients $w_{\mathrm{data}}$, $w_{\mathrm{phys}}$, $w_{\mathrm{IC}}$, and $w_{\mathrm{BC}}$ control the relative contributions of the different constraints \citep{raissi2019physics,cuomo2022}. The particular form and weighting of these terms depend on the physical problem and the manner in which the constraints are imposed; some initial or boundary conditions may instead be enforced directly through the network representation.

This formulation is useful for both forward and inverse problems. In a forward problem, the governing equation and physical parameters are known and the objective is to approximate the solution. In inverse problems, observational data can also be used to estimate unknown physical parameters or coefficients in the governing equations. In both settings, PINNs provide a mesh-free function representation and can combine heterogeneous information within the same optimization framework \citep{raissi2019physics,cuomo2022}. These properties have made PINNs attractive for problems in which observations are incomplete but meaningful physical equations are available. 

At the same time, optimization can be difficult because the different loss components may generate gradients of substantially different magnitudes. \citet{wang2021} showed that these gradient imbalances can hinder convergence, while \citet{krishnapriyan2021} demonstrated that PINNs can become increasingly difficult to optimize for convection--reaction--diffusion-dominated problems. Related work on singularly perturbed convection--diffusion--reaction equations has also shown that standard PINNs can struggle to resolve sharp boundary-layer structure and that asymptotically informed feature embeddings can improve representation of these stiff solution components \citep{goodarzvand2026asymptotically}. More broadly, benchmark studies indicate that PINN performance is strongly problem-dependent rather than uniformly superior across PDE classes and training strategies \citep{hao2024,grossmann2024}.

These considerations are especially relevant for transport-dominated environmental systems, where solutions may vary over several temporal and spatial scales and where the imposed physical model may itself be incomplete. Therefore, the value of a PINN depends not only on the availability of a governing equation but also on the appropriateness of that equation, the information contained in the observations, and the ability of the optimization procedure to balance those sources of information.
Recent work on singularly perturbed convection--diffusion--reaction problems further shows that standard PINNs may fail to resolve sharp solution features unless the network representation is adapted to the underlying asymptotic structure \citep{goodarzvand2026asymptotically}.

% =========================================================
\subsection{Physics-Informed Learning for Atmospheric Systems}
% =========================================================

Atmospheric prediction is a natural setting for physics-informed learning because meteorological fields evolve through physical processes such as advection, vertical motion, thermodynamic transformation, and diabatic heating. At the same time, atmospheric applications are challenging because the dynamics are high-dimensional and observations may be spatially incomplete. Recent studies have therefore explored several ways of incorporating physical knowledge into machine-learning models for atmospheric prediction and reconstruction.

\citet{chen2021} introduced a physics-informed generative neural network for tropospheric temperature prediction using ERA5 reanalysis data, demonstrating the potential value of incorporating physical information into atmospheric thermal prediction. Although their architecture differs from a conventional residual-based PINN, the study is directly relevant to the use of physics-informed learning for temperature forecasting.

Other work has focused on reconstruction from sparse atmospheric observations. \citet{morenoSoto2024} used PINNs with Navier--Stokes-based physical regularization to reconstruct high-resolution wind and pressure fields from sparse weather-station measurements. \citet{eusebi2024} similarly reconstructed two- and three-dimensional tropical-cyclone wind and pressure fields from sparse information and demonstrated the approach using observations associated with Hurricane Ida. Their three-dimensional formulation used pressure as the vertical coordinate, providing a particularly relevant example of pressure-coordinate physics-informed modeling. More recently, \citet{liu2026} proposed a region-growing PINN for wind-field reconstruction under sparse observations, progressively expanding the training domain from data-dense to more sparsely observed regions and using adaptive weighting to improve training stability. These studies demonstrate that physical constraints can help reconstruct atmospheric structure when observations are incomplete, although their primary objective is reconstruction rather than prediction of future thermal states.

Physics-informed atmospheric modeling also extends beyond conventional PINNs. \citet{verma2024} introduced ClimODE, a continuous-time neural model that incorporates atmospheric transport and conservation principles into weather prediction. This work provides broader evidence that explicit transport physics can be useful in learned atmospheric evolution even when the architecture is not formulated as a PDE-residual PINN. More broadly, \citet{waqasKim2026} identify growing use of PINNs in weather and hydrological systems while noting that high-dimensional atmospheric forecasting, sparse field-scale validation, computational scalability, and operational deployment remain important challenges.

Taken together, the literature shows that physics-informed learning has been applied to atmospheric temperature prediction, sparse weather-field reconstruction, pressure-coordinate modeling, and transport-informed forecasting. However, these studies address different tasks and modeling formulations. In particular, successful reconstruction of an atmospheric state from incomplete observations does not necessarily establish that a physics-informed model will improve future thermal-state prediction beyond strong data-driven alternatives. This distinction motivates the sparse-observation and forecasting questions considered in the following section.

% =========================================================
\subsection{PINNs under Sparse and Real-World Observations}
% =========================================================

Recent evidence suggests that the main challenge is no longer simply whether physical constraints can be incorporated into neural models, but when they provide useful predictive information under realistic data limitations. The systematic review by \citet{waqasKim2026} distinguishes relatively established PINN applications in areas such as groundwater and shallow-water modeling from less mature applications in high-dimensional atmospheric forecasting, and identifies sparse field-scale validation, training instability, loss-weight sensitivity, computational cost, scalability, and uncertainty quantification among the remaining challenges.

A particularly important distinction is between atmospheric reconstruction and forecasting. In reconstruction problems, some observations of the atmospheric state being estimated are available at the same time, whereas forecasting requires propagation from an observed initial state into a future period. In the present study, future target temperature is withheld during forecasting, so any benefit of the physics constraint must arise from its contribution to the modeled evolution rather than from concurrent temperature information at the prediction time. The reviewed literature provides comparatively limited evidence on how this benefit changes jointly with forecast lead time and temperature-observation density, whether the same forecasting behavior transfers across regions under a fixed modeling protocol, and how the suitability of a fixed pressure-level representation changes with terrain. The present study addresses these questions through controlled variation of forecast horizon and temperature availability, fixed-protocol cross-region evaluation, and a terrain-related applicability test using real ERA5 atmospheric fields and a synthetic sparse-observation network.

% =========================================================
\section{Methodology and Study Design}
\label{sec:methodology}

\subsection{Data Source and Experimental Design}
\label{sec:data-design}

The experiments use hourly ERA5 reanalysis produced by the European Centre for Medium-Range Weather Forecasts (ECMWF). ERA5 combines observations with the ECMWF Integrated Forecasting System through data assimilation and provides spatially and temporally complete atmospheric fields suitable for controlled retrospective forecasting experiments \citep{hersbach2020}. The pressure-level variables used in this study are air temperature \(T\), zonal wind \(u\), meridional wind \(v\), pressure vertical velocity \(\omega\), and specific humidity \(q\). Surface pressure \(p_s\) is used to determine whether individual pressure levels lie above the local terrain.

The forecasting experiments use three pressure levels, 700, 850, and 925~hPa, which were fixed during model development and retained unchanged in all subsequent experiments. For each forecast origin, 12~h of preceding temperature information are available, and predictions are evaluated at one-, two-, and three-hour lead times. Model training is performed over a geographical domain extending \(4^\circ\) beyond the inner region on which forecast accuracy is evaluated.

The experimental sequence was designed to distinguish model development from subsequent tests of temporal replication, observation sparsity, regional transfer, and terrain-related applicability. Three states, Oklahoma, Alabama, and Montana, were selected to serve complementary experimental roles: Oklahoma for model development and temporal replication, Alabama for cross-region evaluation under the fixed protocol, and Montana as a contrasting higher-terrain region for assessing the applicability of the fixed pressure-level formulation. 

The selected analysis periods were August 14--17, 2026, for
the Oklahoma development experiment; August 18--21, 2026, for the Oklahoma
replication and observation-density experiments; July 26--29, 2026, for Alabama;
and July 24--27, 2026, for Montana. The selected windows coincide with documented
periods of substantial heat, and within each period, four forecast origins were
evaluated, as summarized in Table~\ref{tab:experimental-design}.

\begin{table}[t]
\centering
\caption{Summary of the experimental design across study regions.}
\label{tab:experimental-design}
\footnotesize
\setlength{\tabcolsep}{3pt}
\renewcommand{\arraystretch}{1.2}

\begin{tabular}{
>{\raggedright\arraybackslash}p{2.6cm}
>{\raggedright\arraybackslash}p{2.9cm}
>{\raggedright\arraybackslash}p{3.4cm}
>{\raggedright\arraybackslash}p{1.9cm}
>{\raggedright\arraybackslash}p{3.5cm}
}
\toprule
\textbf{Region / experiment} &
\textbf{Inner evaluation domain} &
\textbf{Forecast origins (UTC)} &
\textbf{Temperature setting} &
\textbf{Experimental role} \\
\midrule

Oklahoma development &
\(33.5^\circ\)--\(37.0^\circ\) N,
\(100.0^\circ\)--\(94.5^\circ\) W &
Aug.\ 15 12:00, 18:00; Aug.\ 16 12:00, 18:00 &
Dense &
Model development \\

Oklahoma chronological replication &
Same &
Aug.\ 19 12:00, 18:00; Aug.\ 20 12:00, 18:00 &
Dense &
Later-period replication under the fixed protocol \\

Oklahoma density study &
Same &
Same four origins &
25\%, 10\%, 5\%; dense reference &
Observation-density \(\times\) forecast-horizon analysis \\

Alabama &
\(31.0^\circ\)--\(34.5^\circ\) N,
\(88.5^\circ\)--\(85.0^\circ\) W &
Jul.\ 27 12:00, 18:00; Jul.\ 28 12:00, 18:00 &
10\% &
Fixed-protocol cross-region evaluation and layout robustness \\

Montana &
\(45.0^\circ\)--\(47.5^\circ\) N,
\(109.0^\circ\)--\(104.5^\circ\) W &
Jul.\ 25 12:00, 18:00; Jul.\ 26 12:00, 18:00 &
Nominal 10\% &
Terrain/applicability stress test \\

\bottomrule
\end{tabular}
\end{table}

The scientific protocol, including the pressure levels, PINN architecture, loss construction, training schedule, source parameterization, terrain rule, and baseline definitions, was fixed after the development stage and retained in subsequent experiments. Model weights were nevertheless re-estimated separately for each forecast origin from the information available before that origin.

\subsubsection{Sparse Observation Design}
\label{sec:sparse-observations}

Observation sparsity is applied only to temperature. The retained values are ERA5 temperatures, while the simulated component is the set of horizontal locations at which temperature is treated as observed. We therefore refer to these as virtual observations rather than physical weather stations.

Eligible locations are determined from spatial geometry and terrain validity only; temperature values are not used in sensor placement. Farthest-point sampling is then used to obtain spatially distributed observation networks. For Oklahoma, 1,723 common candidate locations were available. Nested networks containing approximately 5\%, 10\%, and 25\% of these locations were constructed from a common ordering, corresponding to 87, 173, and 431 locations, respectively:

\[
5\% \subset 10\% \subset 25\%.
\]

This design changes observation density while retaining a consistent spatial layout.

In sparse experiments, only the retained locations provide temperature information during the historical window and at forecast initialization. Initial temperatures are reconstructed horizontally by linear interpolation, with nearest-neighbor interpolation used where necessary, and then converted to potential temperature. Hidden ERA5 temperatures at unobserved locations are therefore not used to initialize the sparse PINN.

Alabama uses the same 10\% rule, retaining 213 of 2,125 eligible locations. Robustness to observation layout is examined using three farthest-point initialization seeds while keeping all other settings fixed. Montana follows the same nominal 10\% rule, although terrain substantially reduces the number of eligible locations; its realized observation fraction is therefore reported separately.
```

\subsection{Physical Model}
\label{sec:physical-model}

\subsubsection{Potential Temperature and Governing Equation}
\label{sec:potential-temperature}

The PINN predicts potential temperature rather than raw air temperature. Potential temperature is the temperature that a dry air parcel would attain if brought adiabatically to a reference pressure and is defined as

\begin{equation}
\theta
=
T\left(\frac{p_0}{p}\right)^\kappa,
\qquad
\kappa=\frac{R_d}{c_p},
\label{eq:potential-temperature}
\end{equation}

where \(p_0=1000\)~hPa, \(R_d=287.05~\mathrm{J\,kg^{-1}\,K^{-1}}\), and
\(c_p=1004~\mathrm{J\,kg^{-1}\,K^{-1}}\), giving
\(\kappa\approx0.286\) \citep{holton2013}. Potential temperature is used because dry adiabatic temperature changes associated with vertical pressure changes are incorporated into the transformation, providing a more natural prognostic variable for pressure-coordinate thermodynamic transport.

The physical constraint imposed on the PINN is

\begin{equation}
\frac{\partial\theta}{\partial t}
+
u\frac{\partial\theta}{\partial x}
+
v\frac{\partial\theta}{\partial y}
+
\omega\frac{\partial\theta}{\partial p}
=
Q_\theta ,
\label{eq:thermodynamic-pde}
\end{equation}

where \(x\) and \(y\) are local horizontal coordinates, \(p\) is pressure, \(t\) is time, \(\omega=Dp/Dt\) is pressure vertical velocity, and \(Q_\theta\) represents unresolved diabatic thermal tendencies expressed as potential-temperature tendency. The first term represents local temporal change. The second and third terms represent horizontal advection by the resolved wind, and the fourth represents vertical transport in pressure coordinates. Equation~\eqref{eq:thermodynamic-pde} is therefore an advection--source equation, a reduced thermodynamic transport model, and unresolved mixing is consequently represented indirectly through the data fit and empirical source closure discussed in Section~3.2.2.

For numerical consistency, the physical coordinates are expressed in hours, kilometers, hPa, and kelvin. The ERA5 horizontal winds are converted from \(\mathrm{m\,s^{-1}}\) to \(\mathrm{km\,h^{-1}}\),

\begin{equation*}
u_{\mathrm{km/h}}=3.6\,u_{\mathrm{m/s}},
\qquad
v_{\mathrm{km/h}}=3.6\,v_{\mathrm{m/s}},
\end{equation*}

and pressure vertical velocity is converted from \(\mathrm{Pa\,s^{-1}}\) to \(\mathrm{hPa\,h^{-1}}\),

\begin{equation*}
\omega_{\mathrm{hPa/h}}
=
36\,\omega_{\mathrm{Pa/s}}.
\end{equation*}

The latitude--longitude grid was converted to approximate local Cartesian coordinates using

\begin{equation}
y = 111(\phi-\phi_0),
\qquad
x = 111\cos(\phi_0)(\lambda-\lambda_0),
\label{eq:local-cartesian}
\end{equation}

where \(\phi\) and \(\lambda\) denote latitude and longitude in degrees, \(\phi_0\) and \(\lambda_0\) are the reference latitude and longitude of the regional domain, and \(x\) and \(y\) are horizontal distances in kilometers. The factor \(111\) approximates the kilometers per degree of latitude, while the cosine term accounts for the latitude dependence of longitudinal distance. Because the study domains are regionally limited, this local planar approximation was used instead of a full spherical-coordinate treatment.

\subsubsection{Empirical Diabatic-Source Closure}
\label{sec:diabatic-closure}

The advection--source equation in Equation~\eqref{eq:thermodynamic-pde} alone does not account for diabatic processes such as latent heating, radiation, turbulent exchange, and other unresolved tendencies and requires a diabatic potential-temperature tendency \(Q_\theta\), which is not directly observable in ERA5. The estimation of unknown source terms from observed state variables is a related inverse problem that also arises in classical heat-transfer models \cite{shivanian2025source}. We therefore use a low-dimensional empirical closure,

\begin{equation}
Q_\theta
=
\alpha_L Q_{\rm latent}
+
b_0
+
b_p p^*
+
b_s\sin\left(\frac{2\pi h_{\rm UTC}}{24}\right)
+
b_c\cos\left(\frac{2\pi h_{\rm UTC}}{24}\right),
\label{eq:source-closure}
\end{equation}

where

\begin{equation}
p^*=\frac{p-850}{150}.
\label{eq:normalized-pressure}
\end{equation}

Here, \(Q_{\rm latent}\) is a humidity-based proxy for latent-heating effects, \(\alpha_L\) is its fitted scaling coefficient, \(b_0\) represents a constant residual tendency, and \(b_p p^*\) allows the residual tendency to vary with pressure. The normalized pressure coordinate \(p^*\) is centered on 850~hPa and scaled by 150~hPa, so that
\[
p^* \in \{-1,\,0,\,0.5\}
\]
at the three pressure levels used in this study. The terms
\[
b_s\sin\left(\frac{2\pi h_{\rm UTC}}{24}\right)
+
b_c\cos\left(\frac{2\pi h_{\rm UTC}}{24}\right)
\]
represent a diurnal harmonic associated with the daily heating--cooling cycle.

The humidity-based term is constructed from the material derivative of ERA5 specific humidity,

\begin{equation}
\frac{Dq}{Dt}
=
\frac{\partial q}{\partial t}
+
u\frac{\partial q}{\partial x}
+
v\frac{\partial q}{\partial y}
+
\omega\frac{\partial q}{\partial p},
\label{eq:humidity-material-derivative}
\end{equation}

A latent-heating proxy in temperature-tendency units is then defined as

\begin{equation}
Q_{\mathrm{latent},T}
=
-\frac{L_v}{c_p}\frac{Dq}{Dt},
\label{eq:latent-temperature-tendency}
\end{equation}

where \(L_v=2.5\times10^6~\mathrm{J\,kg^{-1}}\) is the latent heat of vaporization. This quantity is transformed to potential-temperature tendency as

\begin{equation}
Q_{\rm latent}
=
-\frac{L_v}{c_p}
\frac{Dq}{Dt}
\left(\frac{p_0}{p}\right)^\kappa ,
\label{eq:latent-potential-temperature-tendency}
\end{equation}

where \(L_v=2.5\times10^6~\mathrm{J\,kg^{-1}}\). This quantity is used as an empirical proxy rather than as a direct measurement of latent heating, because changes in specific humidity can also arise from transport, mixing, moisture sources and sinks, and reanalysis adjustments.

The five closure coefficients
\[
(\alpha_L,b_0,b_p,b_s,b_c)
\]
are estimated jointly with the neural-network parameters using only the historical training period. Their ranges are constrained to

\begin{equation}
0<\alpha_L<2,
\qquad
|b_0|,|b_p|\le0.30~\mathrm{K\,h^{-1}},
\qquad
|b_s|,|b_c|\le0.50~\mathrm{K\,h^{-1}}.
\label{eq:closure-bounds}
\end{equation}

These bounds are regularizing design constraints rather than fundamental physical limits. The positivity constraint on \(\alpha_L\) preserves the sign convention of the humidity-derived proxy, while the upper limit prevents arbitrary amplification of that proxy. Bounds on the remaining coefficients restrict the empirical closure from absorbing arbitrarily large unresolved tendencies. The same parameterization and bounds are retained in all subsequent experiments.

After the historical training stage, all five coefficients are frozen before future-domain physics training begins. Thus, the closure parameters cannot adapt to future temperature errors or evaluation targets. Future ERA5 \(q\), along with \(u\), \(v\), and \(\omega\), is still supplied as meteorological forcing when evaluating \(Q_{\rm latent}\) in the future domain; this is part of the conditional hindcast setting described in Section~3.4.

\subsection{PINN Architecture and Physics-Informed Objective}
\label{sec:pinn-architecture}

\subsubsection{Network Representation and Initial-State Constraint}
\label{sec:network-representation}

The PINN represents potential temperature as a function of four physical coordinates,

\[
(x,y,p,\tau)\longmapsto
\widehat{\theta}(x,y,p,\tau),
\]

where \(\tau\) denotes time relative to the forecast origin, with \(\tau=0\) the initialization time. The neural component is a fully connected multilayer perceptron with four hidden layers, 64 neurons per hidden layer, hyperbolic-tangent activation functions, and one scalar output. Two neural-network initialization seeds are used to assess sensitivity to random initialization.

Rather than treating the initial condition as an additional constraint that the network must learn approximately, it is imposed exactly through the model representation,

\begin{equation}
\widehat{\theta}_{\boldsymbol{\vartheta}}
(x,y,p,\tau)
=
\theta_0(x,y,p)
+
\frac{\tau}{H}s_d
N_{\boldsymbol{\vartheta}}
(\widetilde{x},\widetilde{y},\widetilde{p},\widetilde{\tau}),
\label{eq:anchored-representation}
\end{equation}

where \(H=12\)~h, \(N_{\boldsymbol{\vartheta}}\) is the trainable neural network, \(s_d\) is a characteristic temperature scale estimated from historical observations, and the tilded variables denote scaled network inputs. The horizontal coordinates and pressure are mapped to \([-1,1]\), while relative time is scaled by \(H\). Because the neural correction vanishes at \(\tau=0\),

\begin{equation}
\widehat{\theta}_{\boldsymbol{\vartheta}}(x,y,p,0)
=
\theta_0(x,y,p),
\label{eq:exact-initial-state}
\end{equation}

so the initial state is satisfied exactly through construction rather than a soft initial-condition penalty. This construction follows the general hard-constraint or trial-function principle used to enforce known conditions exactly in physics-informed neural models \citep{sukumar2022}.

For dense experiments, \(\theta_0\) is constructed from the complete terrain-valid ERA5 temperature field at the forecast origin. For sparse experiments, it is reconstructed exclusively from retained virtual observations as described in Section~3.5. The initial field is represented differentiably in the horizontal and pressure coordinates so that derivatives of the complete model can be obtained through automatic differentiation.

Meteorological forcing is not supplied as a direct neural-network input. Instead, \(u\), \(v\), \(\omega\), and the source closure enter through the physical residual

\begin{equation}
r_{\boldsymbol{\vartheta}}
=
\frac{\partial\widehat{\theta}}{\partial\tau}
+
u\frac{\partial\widehat{\theta}}{\partial x}
+
v\frac{\partial\widehat{\theta}}{\partial y}
+
\omega\frac{\partial\widehat{\theta}}{\partial p}
-
Q_\theta .
\label{eq:pinn-residual}
\end{equation}

Automatic differentiation is used to evaluate the derivatives of \(\widehat{\theta}\), following the standard PINN formulation. A manufactured linear-field test is used to verify that the implementation of Equation~\eqref{eq:pinn-residual} produces a near-zero residual when the exact derivatives and source satisfy the prescribed balance.

% =========================================================

\subsubsection{Loss Formulation}
\label{sec:loss-formulation}

The temperature-data error and PDE residual have different units and numerical scales. To prevent either term from dominating the optimization simply because of its magnitude, both are normalized using scales calculated from the historical temperature data. The temperature scale is defined as

\begin{equation}
s_d=
\max\left[
\operatorname{SD}\!\left(\theta_{\rm hist}-\theta_0\right),
\,0.5~\mathrm{K}
\right],
\label{eq:data-scale}
\end{equation}

where \(s_d\) represents the typical magnitude of historical potential-temperature departures from the initialized state. The physics scale is

\begin{equation}
s_f=
\max\left[
\operatorname{SD}\!\left(\Delta_{1{\rm h}}\theta_{\rm hist}\right),
\,0.10~\mathrm{K\,h^{-1}}
\right],
\label{eq:physics-scale}
\end{equation}

where \(s_f\) represents the typical magnitude of hourly potential-temperature changes. The lower limits of \(0.5\)~K and \(0.10~\mathrm{K\,h^{-1}}\) were fixed during model development as numerical safeguards to prevent unusually small variability from producing very large normalized losses; they are not physical thresholds and were kept unchanged across all subsequent experiments. In sparse experiments, both scales are calculated only from the retained virtual observations.

The normalized data and physics losses are

\begin{equation}
\mathcal L_{\rm data}
=
\operatorname{MSE}
\left[
\frac{\widehat{\theta}-\theta_{\rm obs}}{s_d}
\right],
\label{eq:data-loss}
\end{equation}

and

\begin{equation}
\mathcal L_{\rm phys}
=
\operatorname{MSE}
\left[
\frac{r_{\boldsymbol{\vartheta}}}{s_f}
\right].
\label{eq:physics-loss}
\end{equation}

A weak outer-boundary term is also used during future-time training,

\begin{equation}
\mathcal L_{\rm bnd}
=
\operatorname{MSE}
\left[
\frac{
\widehat{\theta}(x_b,y_b,p_b,\tau)
-\theta_0(x_b,y_b,p_b)}
{s_d}
\right],
\label{eq:boundary-loss}
\end{equation}

which discourages large changes at the artificial outer boundary without supplying future temperature observations there. No separate initial-condition loss is required because the representation in Equation~\eqref{eq:anchored-representation} satisfies the initial state exactly.

\subsection{Physics Collocation and Staged Training Procedure}
\label{sec:collocation-training}

Physics collocation points are sampled continuously in \(x\), \(y\), \(p\), and \(\tau\), rather than only at ERA5 grid locations. Pressure is sampled continuously between 700 and 925~hPa, while \(u\), \(v\), \(\omega\), and the humidity-derived source fields are interpolated from the ERA5 forcing fields.

A point is considered terrain-valid only when

\begin{equation}
p_s(x,y,t)\ge p+20~\mathrm{hPa}.
\label{eq:terrain-validity}
\end{equation}

The additional 20-hPa margin was introduced during model development to exclude pressure surfaces immediately adjacent to terrain, where surface intersection and interpolation become ambiguous. It is a fixed modeling safeguard rather than a universal atmospheric threshold and was retained across all experiments. The inner evaluation domain is further surrounded by a \(4^\circ\) buffer to reduce sensitivity to the approximate outer-boundary treatment.

Training is performed in two phases.

During the \textbf{historical source-learning phase}, only the 12~h preceding the forecast origin are used,

\[
-12\le\tau<0.
\]

Both the neural-network parameters and the five source-closure coefficients
\[
(\alpha_L,b_0,b_p,b_s,b_c)
\]
are estimated during this phase. Because \(b_0\), \(b_p\), \(b_s\), and \(b_c\) represent empirical corrections to unresolved thermal tendencies, a weak quadratic regularization is applied,

\begin{equation}
\mathcal L_{\rm src}
=
\left(\frac{b_0}{0.30}\right)^2+
\left(\frac{b_p}{0.30}\right)^2+
\left(\frac{b_s}{0.50}\right)^2+
\left(\frac{b_c}{0.50}\right)^2.
\label{eq:source-regularization}
\end{equation}

This term does not represent an additional physical constraint; it simply discourages unnecessarily large empirical source corrections while those coefficients are being estimated.

The historical-stage objective is

\begin{equation}
\mathcal L_A
=
w_{\rm data}\mathcal L_{\rm data}
+
w_{\rm phys}\mathcal L_{\rm phys}
+
w_{\rm src}\mathcal L_{\rm src},
\label{eq:historical-objective}
\end{equation}

with

\[
w_{\rm data}=1,\qquad
w_{\rm phys}=1,\qquad
w_{\rm src}=10^{-3}.
\]

After 1,500 Adam steps, all five source-closure coefficients are frozen. The source term \(Q_\theta\) remains part of the governing equation, but its fitted coefficients are no longer updated.

During the \textbf{future-physics phase}, the constrained time domain is expanded progressively through one, two, and three hours,

\[
0<\tau\le1
\;\rightarrow\;
0<\tau\le2
\;\rightarrow\;
0<\tau\le3~\mathrm{h}.
\]

At stage \(h\in\{1,2,3\}\), the physics contribution is

\begin{equation}
\mathcal L_{\rm phys}^{(h)}
=
\frac{1}{2}
\left[
\operatorname{MSE}
\left(\frac{r_{\rm past}}{s_f}\right)
+
\operatorname{MSE}
\left(\frac{r_{0<\tau\le h}}{s_f}\right)
\right],
\label{eq:staged-physics-loss}
\end{equation}

and the optimization objective becomes

\begin{equation}
\mathcal L_h
=
w_{\rm data}\mathcal L_{\rm data}
+
w_{\rm phys}\mathcal L_{\rm phys}^{(h)}
+
w_{\rm bnd}\mathcal L_{\rm bnd},
\label{eq:future-objective}
\end{equation}

with

\[
w_{\rm data}=1,\qquad
w_{\rm phys}=1,\qquad
w_{\rm bnd}=0.05.
\]

The source-regularization term is no longer needed because the source coefficients are fixed; however, the fitted source closure \(Q_\theta\) continues to enter the PDE residual. Each future stage uses 1,000 additional Adam steps. The progressive one-, two-, and three-hour expansion introduces future physics gradually rather than imposing the complete three-hour space--time problem at once, consistent with broader curriculum and causality-aware strategies for time-dependent PINNs \citep{krishnapriyan2021,wang2024}.

Future ERA5 \(u\), \(v\), \(\omega\), and \(q\) are supplied during these stages, while future temperature is withheld until evaluation. The experiments therefore represent conditional retrospective forecasts rather than autonomous operational forecasts.

\subsection{Comparative Evaluation Framework}
\subsubsection{Baseline Models and Comparison Strategy}
\label{sec:baselines}

The PINN is compared with four baselines, each chosen to rule out a specific alternative explanation for any PINN advantage.

The \textbf{persistence baseline}, often a strong baseline at the shortest lead times, assumes that the initialized temperature field remains unchanged over the forecast period,

\begin{equation}
\widehat{\theta}_{\rm pers}
=
\theta_0.
\label{eq:persistence-baseline}
\end{equation}

The \textbf{recent-trend baseline} estimates a local temporal slope from the four most recent hourly temperature values and extrapolates that trend forward,

\begin{equation}
\widehat{\theta}_{\rm trend}
=
\theta_0+\widehat{s}\tau.
\label{eq:trend-baseline}
\end{equation}

Under sparse conditions, both the initial field and temporal trends are constructed only from retained virtual observations.

The \textbf{coordinate-only neural network} uses the same four-layer, 64-unit tanh architecture and exact initial-state anchoring as the PINN but is trained only from historical temperature data.

The \textbf{forcing-aware neural network} uses the same anchored architecture and additionally receives \(u\), \(v\), \(\omega\), \(q\), \(Q_{\rm latent}\), and the diurnal sine/cosine terms, but it does not use a PDE-residual loss. During forecasting, it receives the same category of future meteorological forcing as the PINN. This provides a direct comparison between supplying meteorological predictors and explicitly enforcing the governing equation.

Both neural baselines use the same available temperature information and overall 4,500-step optimization budget as the PINN.

For a conservative comparison, performance is also evaluated against a best-of-baselines oracle envelope,

\begin{equation}
{\rm RMSE}_{\rm best}
=
\min
\left\{
{\rm RMSE}_{\rm pers},
{\rm RMSE}_{\rm trend},
{\rm RMSE}_{\rm coordNN},
{\rm RMSE}_{\rm forcingNN}
\right\}.
\label{eq:oracle-baseline}
\end{equation}

Because the lowest-error baseline is selected after the forecast truth is known, this envelope is intentionally stronger than any individual baseline but is not itself a deployable forecasting method. A PINN win is recorded when its RMSE is lower than the oracle baseline for the corresponding evaluated case.

% =========================================================

\subsubsection{Evaluation Protocol}

Predictions are evaluated at one-, two-, and three-hour leads over terrain-valid points in the inner evaluation domain. Predicted potential temperature is converted back to air temperature using

\begin{equation}
\widehat{T}
=
\widehat{\theta}
\left(\frac{p}{p_0}\right)^\kappa .
\label{eq:temperature-reconstruction}
\end{equation}

The primary forecast-error metric is root-mean-square error,

\begin{equation}
{\rm RMSE}
=
\sqrt{
\frac{1}{N}
\sum_{i=1}^{N}
(\widehat{T}_i-T_i)^2
},
\label{eq:rmse}
\end{equation}

reported in kelvin. Mean absolute error and correlation are retained as secondary diagnostics.

Relative improvement over the oracle baseline envelope is calculated as

\begin{equation}
I=
100
\frac{
{\rm RMSE}_{\rm best}
-
{\rm RMSE}_{\rm PINN}}
{{\rm RMSE}_{\rm best}}.
\label{eq:relative-improvement}
\end{equation}

Positive values indicate lower RMSE for the PINN. A case is counted as a PINN win when

\[
{\rm RMSE}_{\rm PINN}<{\rm RMSE}_{\rm best}.
\]

For the Oklahoma development and chronological-replication experiments, a practical screening criterion was defined as at least 5\% mean RMSE improvement relative to the oracle baseline envelope, a PINN win in at least 67\% of origin--seed--lead combinations (superiority in at least two-thirds of all evaluated cases), and a win in at least 50\% of three-hour cases. These values are project-defined decision criteria. The 10\% sparse Oklahoma experiment was evaluated against the same criteria without retrospectively changing them.

For the frozen Alabama evaluation, the three-hour lead was designated as the primary endpoint. The prespecified criterion required at least 5\% mean improvement at three hours and PINN superiority at at least three of the four forecast origins after averaging the two neural initialization seeds. The subsequent layout experiment evaluates whether the same three-hour behavior persists across the three virtual-observation layouts.

Formal pointwise significance testing is not applied because neighboring ERA5 grid cells are spatially correlated and each regional period contains only four forecast origins. Evidence is therefore summarized descriptively through RMSE, percentage improvement, forecast-horizon behavior, origin-level win rates, neural-seed consistency, observation-layout robustness, chronological replication, and cross-region behavior rather than by treating individual grid cells as independent replicates.

\subsection{Implementation and Verification}
\label{sec:implementation-verification}

All models were implemented in Python using PyTorch and trained on NVIDIA T4 GPU-enabled runtimes. Two neural-network initialization seeds were used throughout the primary experiments; additional observation-layout seeds used for the Alabama robustness analysis are described in the experimental design.

Implementation checks included verification of the temperature--potential-temperature transformation and unit conversions, a manufactured-field test of the PDE residual, terrain-validity checks, and confirmation that the learned source coefficients remained fixed after the historical training stage. Future temperature was excluded from all training losses and future physics collocation, and sparse observation placement depended only on geometry and terrain validity; future ERA5 meteorological variables remained intentionally available as forcing under the conditional-hindcast design.

% =========================================================
\section{Results}
% =========================================================

\subsection{Oklahoma Development and Chronological Replication}
\label{sec:oklahoma-development-replication}

We first evaluated the PINN in the Oklahoma development period to determine whether the physics-informed formulation provided additional predictive value over the comparison methods under dense temperature observations. Performance was evaluated over four forecast origins, two neural-network initialization seeds, and three forecast lead times, giving 24 forecast cases per experiment. Here, a forecast case refers to one origin--seed--lead combination.

During the development period, the PINN achieved a mean RMSE of 0.526~K, compared with 0.636~K for the strongest baseline selected separately for each forecast case, corresponding to a mean RMSE improvement of 17.3\%. The PINN had lower RMSE than the strongest baseline in 17 of 24 cases (70.8\%) and in 5 of the 8 \(+3\)-h cases (62.5\%).

To examine whether this performance persisted beyond the development dates, the same model formulation and training protocol were applied without methodological retuning to a later Oklahoma period. In this chronological replication, the mean PINN RMSE was 0.641~K compared with 0.781~K for the strongest-baseline envelope, giving a mean improvement of 18.0\%. The PINN outperformed the strongest baseline in 18 of 24 cases (75.0\%) and in 6 of the 8 \(+3\)-h cases (75.0\%). As summarized in Table~\ref{tab:oklahoma-dense}, the overall improvement observed during the development period was maintained in the later Oklahoma period, although the PINN was not the best-performing method in every individual forecast case.

\begin{table}[htbp]
\centering
\caption{Oklahoma development and chronological-replication performance under dense temperature observations.}
\label{tab:oklahoma-dense}
\renewcommand{\arraystretch}{1.25}
\setlength{\tabcolsep}{5pt}

\resizebox{\textwidth}{!}{%
\begin{tabular}{lccccc}
\toprule
\textbf{Experiment} &
\shortstack{\textbf{Mean PINN}\\\textbf{RMSE (K)}} &
\shortstack{\textbf{Mean strongest-}\\\textbf{baseline RMSE (K)}} &
\shortstack{\textbf{Mean RMSE}\\\textbf{improvement (\%)}} &
\shortstack{\textbf{Overall PINN}\\\textbf{win rate (\%)}} &
\shortstack{\textbf{\(+3\)-h PINN}\\\textbf{win rate (\%)}} \\
\midrule

Oklahoma development
& 0.526
& 0.636
& 17.3
& 70.8
& 62.5 \\

Oklahoma chronological replication
& 0.641
& 0.781
& 18.0
& 75.0
& 75.0 \\

\bottomrule
\end{tabular}%
}

\vspace{2mm}
\begin{minipage}{\textwidth}
\footnotesize
\textit{Note:} A PINN win indicates lower RMSE than the strongest baseline for that forecast case. The strongest baseline is the lowest-RMSE method among persistence, recent-trend extrapolation, the matched coordinate-only neural network, and the matched forcing-aware neural network.
\end{minipage}

\end{table}

\subsection{Forecast-Horizon and Observation-Density Effects}
\label{sec:horizon-density}

After the chronological replication, we examined how the PINN advantage changed with both forecast lead time and the amount of available temperature information. The same Oklahoma replication period and frozen modeling protocol were used, while temperature-observation coverage was varied from the dense reference to nested 25\%, 10\%, and 5\% networks. These corresponded to 431, 173, and 87 virtual observation locations, respectively.

\begin{figure}[htbp]
\centering

\begin{subfigure}[t]{0.49\textwidth}
    \centering
    \includegraphics[width=\linewidth]{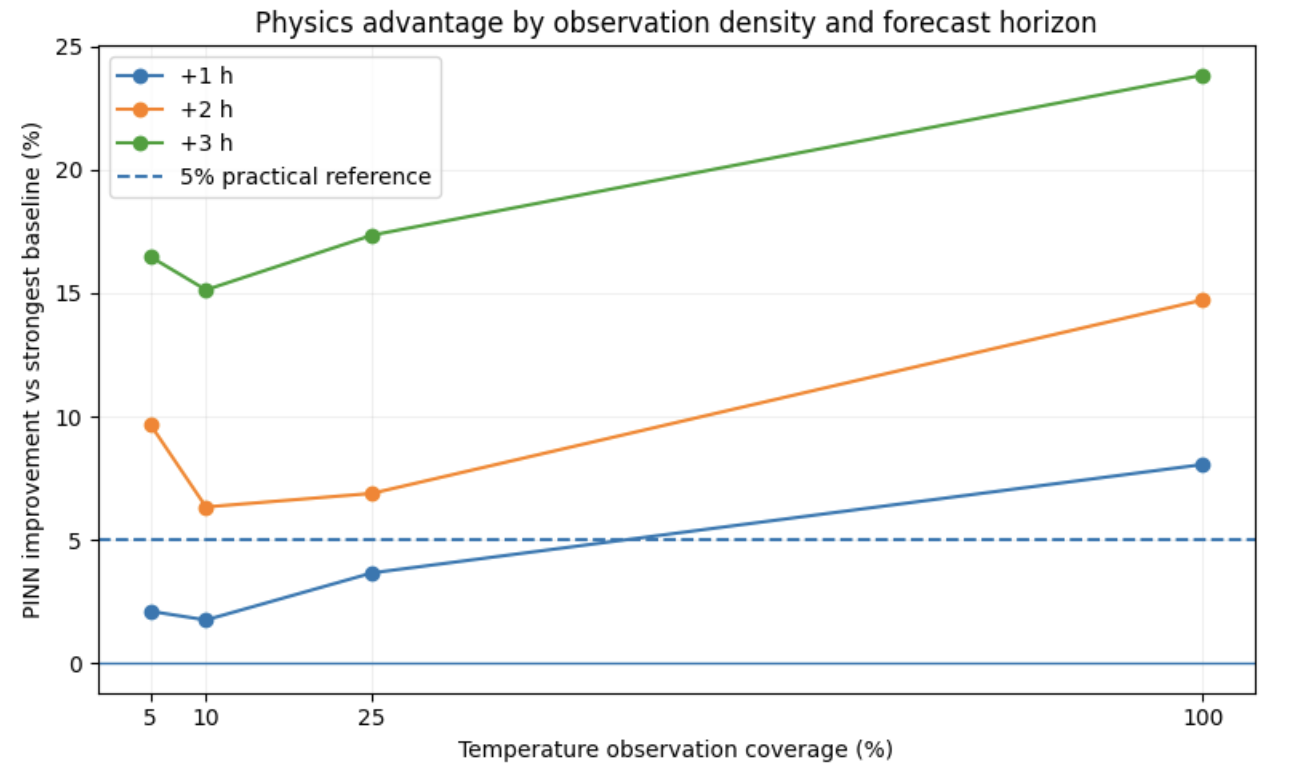}
    \caption{Physics advantage by observation density and forecast horizon. PINN RMSE improvement relative to the strongest baseline at \(+1\), \(+2\), and \(+3\)~h. The dashed line denotes the 5\% practical-improvement reference used in the study.}
    \label{fig:oklahoma-density-a}
\end{subfigure}
\hfill
\begin{subfigure}[t]{0.49\textwidth}
    \centering
    \includegraphics[width=\linewidth]{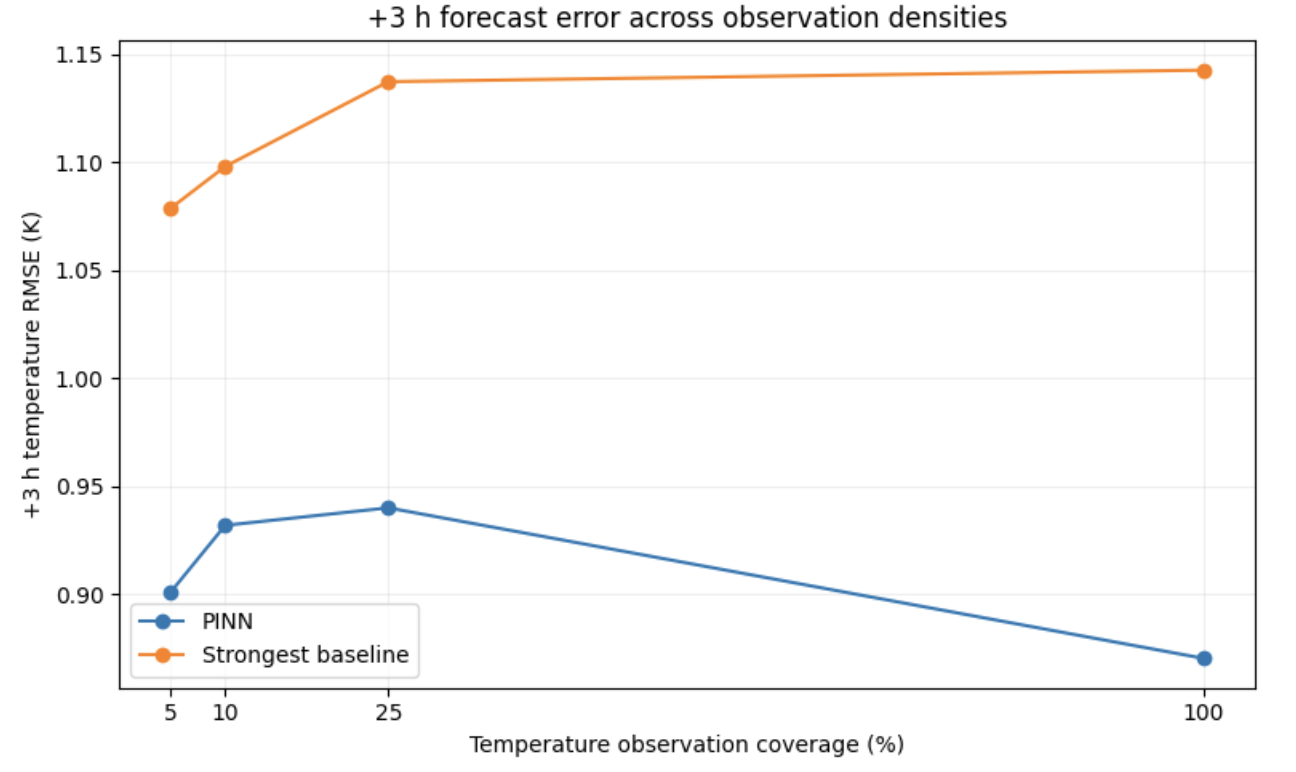}
    \caption{\( +3 \)-h forecast error across observation densities. Mean PINN and strongest-baseline RMSE at the \(+3\)-h forecast horizon.}
    \label{fig:oklahoma-density-b}
\end{subfigure}

\caption{Effect of temperature-observation coverage and forecast horizon on Oklahoma forecast performance.}
\label{fig:oklahoma-density}

\end{figure}

A clear lead-time pattern was observed across all four observation settings (Table~\ref{tab:oklahoma-density} and Fig.~\ref{fig:oklahoma-density}a). Under dense observations, the PINN improvement over the strongest baseline increased from 8.1\% at \(+1\)~h to 14.7\% at \(+2\)~h and 23.8\% at \(+3\)~h. The same pattern remained under sparse observations. At 25\% coverage, the improvement increased from 2.6\% to 5.5\% and 16.9\% across the three lead times; at 10\% coverage, from 1.8\% to 6.3\% and 15.1\%; and at 5\% coverage, from 1.4\% to 9.1\% and 14.6\%. Thus, the PINN provided relatively little additional benefit at the shortest horizon under sparse observations, while its advantage became substantially larger by \(+3\)~h.

The overall mean improvement remained positive at each observation density: 18.0\% for the dense reference and approximately 9--10\% for each sparse setting. At \(+3\)~h, the PINN had lower RMSE than the strongest baseline in 87.5\% of the forecast cases at each sparse density. Figure~\ref{fig:oklahoma-density}b shows that this relative advantage also corresponded to lower absolute RMSE at \(+3\)~h across all four observation settings. The results do not indicate that reducing the number of observations improves the PINN; rather, they show that the \(+3\)-h advantage was retained even as temperature coverage was reduced to 5\%.

The previously defined practical criteria were also applied without modification. The dense, 25\%, and 5\% settings met all three criteria. The 10\% setting achieved a 9.3\% mean improvement and an 87.5\% \(+3\)-h win rate, but its overall win rate of 62.5\% remained below the predefined 67\% criterion.

\begin{table}[htbp]
\centering
\caption{Oklahoma PINN performance by temperature-observation coverage and forecast lead time.}
\label{tab:oklahoma-density}
\renewcommand{\arraystretch}{1.25}
\setlength{\tabcolsep}{4pt}

\resizebox{\textwidth}{!}{%
\begin{tabular}{lcccccc}
\toprule
\shortstack{\textbf{Temperature-observation}\\\textbf{coverage}} &
\shortstack{\textbf{\(+1\)-h}\\\textbf{improvement (\%)}} &
\shortstack{\textbf{\(+2\)-h}\\\textbf{improvement (\%)}} &
\shortstack{\textbf{\(+3\)-h}\\\textbf{improvement (\%)}} &
\shortstack{\textbf{Overall mean}\\\textbf{improvement (\%)}} &
\shortstack{\textbf{Overall PINN}\\\textbf{win rate (\%)}} &
\shortstack{\textbf{\(+3\)-h PINN}\\\textbf{win rate (\%)}} \\
\midrule

Dense (100\%) & 8.1 & 14.7 & 23.8 & 18.0 & 75.0 & 75.0 \\
25\%          & 2.6 & 5.5  & 16.9 & 10.1 & 75.0 & 87.5 \\
10\%          & 1.8 & 6.3  & 15.1 & 9.3  & 62.5 & 87.5 \\
5\%           & 1.4 & 9.1  & 14.6 & 9.8  & 70.8 & 87.5 \\

\bottomrule
\end{tabular}%
}

\vspace{2mm}
\begin{minipage}{\textwidth}
\footnotesize
\textit{Note:} Improvement is the percentage reduction in RMSE relative to the strongest baseline for the corresponding evaluation set. Positive values indicate lower RMSE for the PINN.
\end{minipage}

\end{table}

\subsection{Cross-Region Evaluation and Observation-Layout Robustness in Alabama}
\label{sec:alabama-cross-region}

To address the cross-region component of RQ3, we evaluated whether the Oklahoma findings transferred to Alabama under the same modeling protocol without methodological retuning. The experiment used a 10\% sparse temperature network consisting of 213 of 2,125 eligible virtual observation locations. As in Oklahoma, the models were trained separately for each forecast origin using the preceding 12~h of available data, while the model formulation, architecture, training procedure, pressure levels, and evaluation protocol remained fixed.

The same horizon-dependent pattern observed in Oklahoma was also present in Alabama (Table~\ref{tab:alabama-results}A). The PINN improvement over the strongest baseline increased from 4.6\% at \(+1\)~h to 11.3\% at \(+2\)~h and 19.7\% at \(+3\)~h. At the primary \(+3\)-h endpoint, the PINN achieved a mean RMSE of 0.691~K compared with 0.860~K for the strongest baseline and performed better at three of the four forecast origins after averaging the two neural-network seeds. The predefined Alabama criterion of at least 5\% mean improvement and improvement at three or more of the four forecast origins was therefore satisfied.

Because performance under sparse observations can depend on where the available observations are located, the \(+3\)-h experiment was repeated using two additional 10\% observation layouts while keeping all other settings unchanged. As summarized in Table~\ref{tab:alabama-results}B, the PINN improvement remained positive for all three layouts, ranging from 19.7\% to 24.4\%. It performed better than the strongest baseline at three of four origins for Layout~1 and at all four origins for Layouts~2 and~3. This shows that the Alabama result was not limited to one particular sparse-observation arrangement.

\begin{figure}[htbp]
    \centering
    \includegraphics[width=0.98\textwidth]{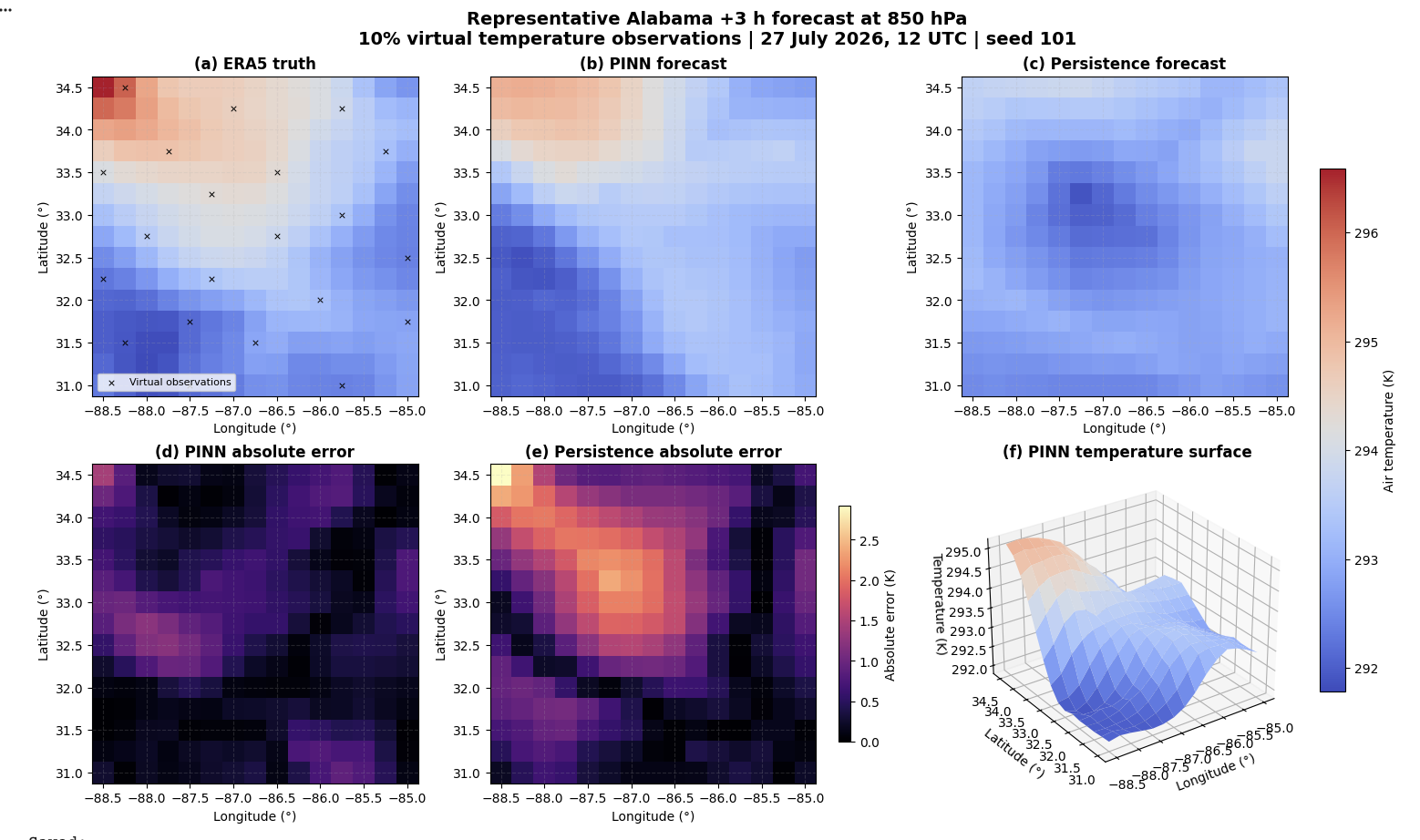}
    
    \caption{Spatial comparison for a representative \(+3\)-h Alabama forecast at 850~hPa under 10\% temperature-observation coverage. The six panels show: (a) ERA5 temperature field and virtual observation locations, (b) PINN forecast, (c) persistence forecast, (d) PINN absolute error, (e) persistence absolute error, and (f) three-dimensional PINN temperature surface. Panels (a)--(c) use a common temperature scale, while panels (d) and (e) use a common absolute-error scale.}
    
    \label{fig:alabama-spatial}
\end{figure}

Figure~\ref{fig:alabama-spatial} provides a spatial example of the \(+3\)-h forecast at 850~hPa. The PINN reproduced the main horizontal temperature pattern more closely than persistence, which was the strongest baseline for the displayed case. The error maps also show generally smaller and more localized errors for the PINN, while persistence produced broader areas of larger error. The displayed example had a PINN RMSE of 0.870~K compared with 1.040~K for persistence, corresponding to a 16.3\% reduction in RMSE. This case is included to illustrate the spatial behavior of the forecast; the Alabama conclusions are based on the full results summarized in Table~\ref{tab:alabama-results}.

\begin{table}[htbp]
\centering
\caption{Alabama cross-region evaluation and robustness to sparse-observation layout.}
\label{tab:alabama-results}
\small
\renewcommand{\arraystretch}{1.2}

\textbf{(A) Performance by forecast lead}

\vspace{1.5mm}

\resizebox{0.96\textwidth}{!}{%
\begin{tabular}{lcccc}
\toprule
\textbf{Forecast lead} &
\shortstack{\textbf{PINN}\\\textbf{RMSE (K)}} &
\shortstack{\textbf{Strongest-baseline}\\\textbf{RMSE (K)}} &
\shortstack{\textbf{RMSE}\\\textbf{improvement (\%)}} &
\shortstack{\textbf{PINN}\\\textbf{win rate (\%)}} \\
\midrule
\( +1 \)~h & 0.373 & 0.391 & 4.6  & 37.5 \\
\( +2 \)~h & 0.573 & 0.646 & 11.3 & 75.0 \\
\( +3 \)~h & 0.691 & 0.860 & 19.7 & 75.0 \\
\bottomrule
\end{tabular}%
}

\vspace{4mm}

\textbf{(B) Robustness across 10\% observation layouts at \(+3\)~h}

\vspace{1.5mm}

\begin{tabular}{lcc}
\toprule
\textbf{Observation layout} &
\shortstack{\textbf{RMSE}\\\textbf{improvement (\%)}} &
\shortstack{\textbf{Origins with}\\\textbf{lower PINN RMSE}} \\
\midrule
Layout 1 & 19.7 & 3/4 \\
Layout 2 & 23.6 & 4/4 \\
Layout 3 & 24.4 & 4/4 \\
\bottomrule
\end{tabular}

\vspace{2mm}

\begin{minipage}{0.96\textwidth}
\footnotesize
\textit{Note:} Layouts 1, 2, and 3 correspond to observation-selection seeds 4242, 13579, and 24680, respectively. The seed values are included only for reproducibility.
\end{minipage}

\end{table}

\subsection{Montana Stress Test and Applicability of the Fixed Pressure-Level Formulation}
\label{sec:montana-stress-test}

\begin{figure}[htbp]
\centering

\begin{subfigure}[t]{0.48\textwidth}
    \centering
    \includegraphics[width=\linewidth]{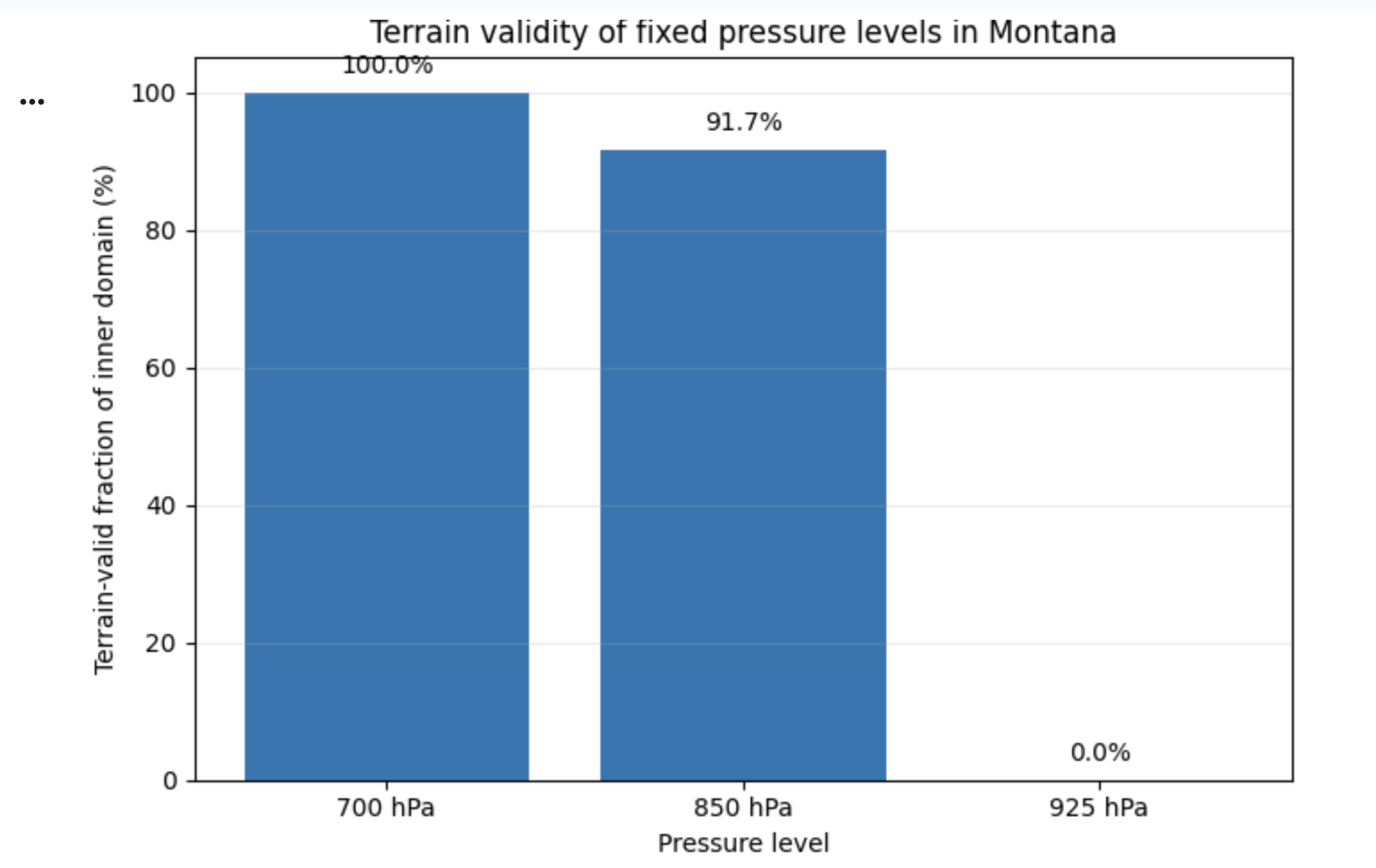}
    \caption{Pressure-level applicability over Montana terrain: fraction of terrain-valid inner-domain grid cells at 700, 850, and 925~hPa.}
    \label{fig:montana-terrain}
\end{subfigure}
\hfill
\begin{subfigure}[t]{0.48\textwidth}
    \centering
    \includegraphics[width=\linewidth]{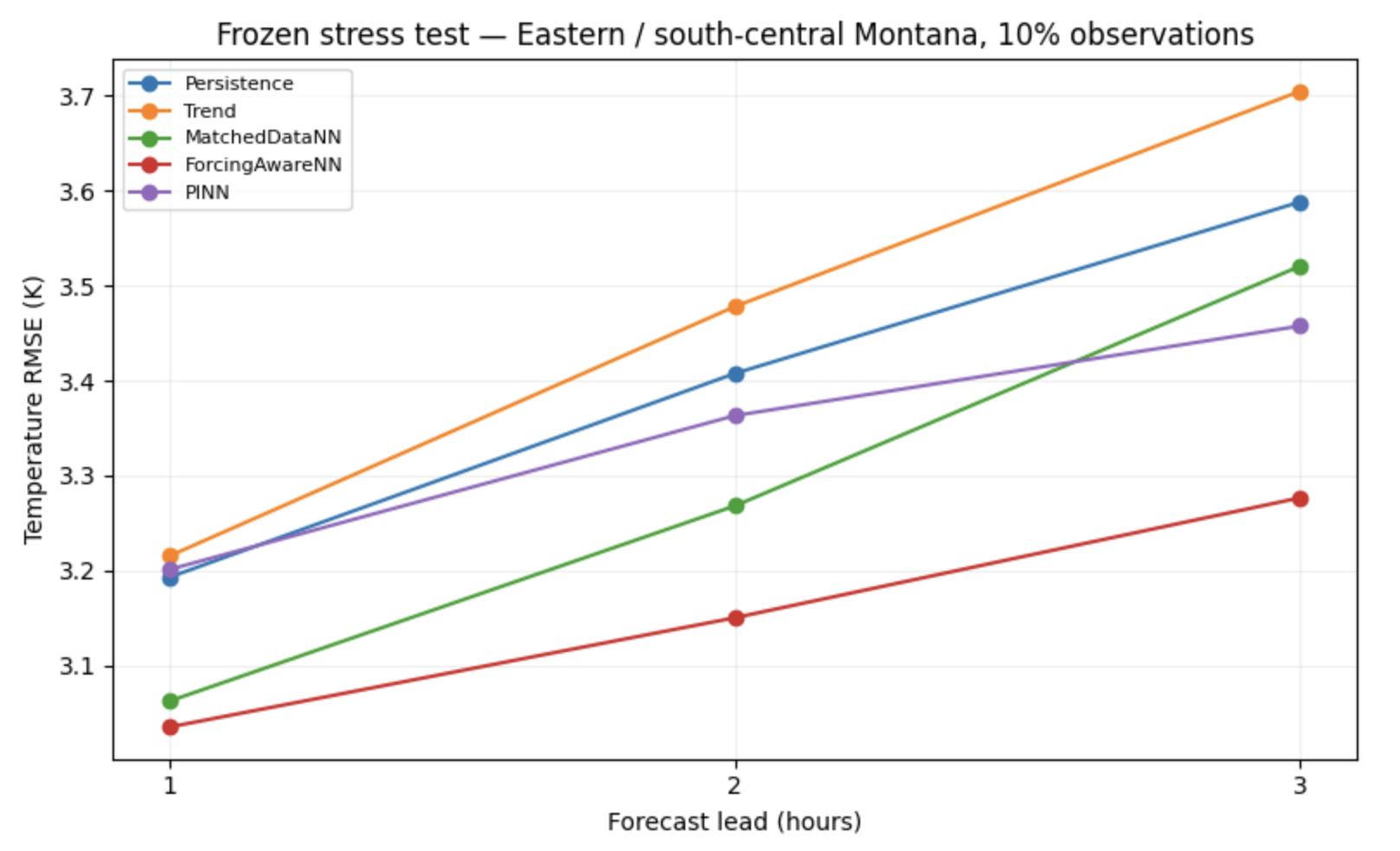}
    \caption{Frozen stress test in eastern/south-central Montana under the nominal 10\% observation setting: mean RMSE of the PINN and comparison methods across \(+1\), \(+2\), and \(+3\)~h forecast horizons.}
    \label{fig:montana-rmse}
\end{subfigure}

\caption{Terrain applicability and forecast performance in the Montana stress test.}
\label{fig:montana-stress}

\end{figure}

To examine the physical-applicability component of RQ3, we applied the same frozen pressure-level formulation in eastern/south-central Montana, where higher terrain provides a more demanding test of the fixed 700-, 850-, and 925-hPa representation. Unlike Oklahoma and Alabama, the Montana domain showed a substantial terrain constraint. Within the inner evaluation domain, the 700-hPa level was valid at all four forecast origins, while 850-hPa validity ranged from 91.4\% to 92.3\%, with a mean of 91.7\%. The 925-hPa level was not terrain-valid anywhere in the inner domain under the prescribed surface-pressure margin (Fig.~\ref{fig:montana-stress}a). 

Sensor selection was performed over the larger $4^\circ$-buffered model domain, where 67 locations remained valid at all three pressure levels across the required histories. The pre-specified minimum-location safeguard retained 25 sensors, corresponding to 37.3\% of this eligible pool. Because 925 hPa was unavailable within the inner evaluation region, the Montana forecast metrics were based on the physically valid 700- and 850-hPa evaluation points.

Using the unchanged formulation, the PINN did not retain the positive advantage observed in Oklahoma and Alabama (Table~\ref{tab:montana-stress}; Fig.~\ref{fig:montana-stress}b). Mean PINN RMSE was higher than the strongest-baseline RMSE at each forecast horizon, with improvement values of \(-8.4\%\), \(-14.0\%\), and \(-17.5\%\) at \(+1\), \(+2\), and \(+3\)~h, respectively. At the primary \(+3\)-h endpoint, the PINN RMSE was 3.458~K compared with 2.942~K for the strongest baseline. After averaging the two neural-network seeds within each origin, the PINN achieved lower RMSE at only one of the four forecast origins. The predefined cross-region criterion was therefore not satisfied.

Taken together, the terrain diagnostic and forecast results identify an important applicability limitation of the fixed pressure-level setup. In particular, the complete loss of terrain-valid coverage at 925~hPa means that the same three-level representation used successfully in Oklahoma and Alabama does not transfer cleanly to this higher-terrain region. The Montana results should therefore be interpreted as a terrain-constrained stress test of the frozen formulation, not as a direct observation-density replication of the Oklahoma or Alabama experiments. A terrain-adaptive vertical representation would be needed to evaluate the approach more fully in such regions.

\begin{table}[htbp]
\centering
\caption{Montana frozen stress-test performance by forecast lead.}
\label{tab:montana-stress}
\small
\renewcommand{\arraystretch}{1.2}
\setlength{\tabcolsep}{6pt}

\begin{tabular}{lccc}
\toprule
\textbf{Forecast lead} &
\shortstack{\textbf{Mean PINN}\\\textbf{RMSE (K)}} &
\shortstack{\textbf{Mean strongest-}\\\textbf{baseline RMSE (K)}} &
\shortstack{\textbf{RMSE}\\\textbf{improvement (\%)}} \\
\midrule
\( +1 \)~h & 3.201 & 2.954 & \(-8.4\) \\
\( +2 \)~h & 3.363 & 2.951 & \(-14.0\) \\
\( +3 \)~h & 3.458 & 2.942 & \(-17.5\) \\
\bottomrule
\end{tabular}

\vspace{2mm}

\begin{minipage}{0.92\textwidth}
\footnotesize
\textit{Note:} Negative improvement indicates that the PINN RMSE was higher than the strongest-baseline RMSE.
\end{minipage}

\end{table}

\section{Discussion}
\label{sec:discussion}

The results show that the value of the physics-informed formulation depends strongly on both observation availability and the physical suitability of the prescribed atmospheric representation. In Oklahoma, the PINN advantage persisted across the development period, a later chronological window, and progressively sparser temperature observations. The Alabama experiment provided an additional test under a different regional setting and retained the positive behavior without retuning the scientific configuration. Together, these results indicate that the improvement was not limited to a single forecast origin, time window, or observation layout.

The observation-density experiments provide the clearest evidence for the role of the physical constraint. As temperature observations became more limited, the PINN continued to use the thermodynamic residual to constrain the reconstructed field. This is consistent with the broader motivation for physics-informed neural networks, in which governing equations provide additional structure when observations alone are insufficient \cite{raissi2019physics,cuomo2022,waqasKim2026}. Related atmospheric applications have likewise examined physics-informed learning under sparse observational information \cite{morenoSoto2024,eusebi2024}. The comparison with the forcing-aware neural network is particularly important because both models receive the same meteorological forcing, while only the PINN is constrained by the thermodynamic equation. The observed differences therefore cannot be attributed solely to access to wind, humidity, and related atmospheric inputs.

The Alabama results also suggest that the selected formulation has some degree of regional robustness. The same pressure levels, architecture, loss weights, training budget, and source formulation were retained, and the model was refit only from the observations associated with each new forecast origin. The consistency across multiple observation layouts further reduces the likelihood that the result was produced by a favorable sparse-network geometry. At the same time, the study includes only a small number of forecast origins and regional heat periods, so broader geographic or climatological generalization cannot yet be established.

Montana provides an important counterexample to unrestricted transfer of the same formulation. The 700-hPa level remained fully valid and 850~hPa remained largely available, but 925~hPa was not terrain-valid anywhere within the inner evaluation domain. The common three-level observation network was consequently limited to terrain-compatible locations in the larger buffered domain, and the reported forecast errors were evaluated only at physically valid 700- and 850-hPa points. Under these conditions, the PINN no longer retained the advantage observed in Oklahoma and Alabama. This result indicates that the fixed 700--850--925-hPa representation becomes restrictive in higher-terrain regions and supports the use of terrain-adaptive pressure levels or an alternative vertical coordinate in future extensions. Terrain-following and hybrid pressure coordinates are commonly used in atmospheric modeling specifically to accommodate complex topography and reduce difficulties associated with fixed pressure surfaces near the ground \cite{beck2020hybrid,choi2021hybrid}.

The study also has several limitations. The sparse observation networks were created by withholding ERA5 temperature values, so the experiments evaluate reconstruction of the ERA5 atmospheric state rather than performance against an independent observational dataset \cite{hersbach2020}. In addition, the model uses future ERA5 meteorological forcing during the forecast period, although future temperature is withheld. Because ERA5 is a retrospective reanalysis product rather than an operational forecast, the results therefore represent short-horizon retrospective forecasts under known meteorological forcing, not fully operational forecasts based only on information available at the forecast origin \cite{hersbach2020}. The unresolved diabatic tendency is represented through the empirical source closure used in this study, whose adequacy may vary across regions and weather regimes. Finally, the Montana experiment showed that the fixed 700--850--925-hPa representation is not suitable in all terrain settings.

Future work should evaluate the framework using independent upper-air observations, a larger collection of weather regimes and geographic regions, and terrain-adaptive vertical representations. Additional work on source-term modeling may also improve the treatment of diabatic processes that are currently absorbed into the empirical closure. These extensions would provide a stronger assessment of when physics-informed constraints offer a reliable advantage for short-horizon atmospheric temperature estimation from sparse observations.

\section{Conclusion}
This study evaluated a physics-informed neural network for short-horizon atmospheric temperature reconstruction from sparse observations using a pressure-coordinate thermodynamic constraint. The results showed consistent benefit in Oklahoma across chronological and observation-density tests and retained positive performance under the frozen Alabama protocol, while the Montana experiment exposed a clear limitation of the fixed pressure-level formulation in higher terrain. Together, these findings suggest that physics-informed constraints can provide useful additional structure for sparse atmospheric temperature estimation when the physical representation remains appropriate to the region. Future work should extend the framework to independent observational data, broader weather regimes, and terrain-adaptive vertical representations.

\section*{Code Availability}

The code, executed notebooks, and reproducibility materials supporting this study are publicly available at:
\url{https://github.com/Tannaz-Chegini/pinn-atmospheric-heat-transfer}.

% REFERENCES
% =========================================================

\bibliographystyle{unsrtnat}
\bibliography{references}

\end{document}